%% file: main.tex
\documentclass[letter, 10 pt, conference]{ieeeconf}  

\IEEEoverridecommandlockouts                              

\usepackage{graphics} 
\usepackage{epsfig} 
\usepackage[linesnumbered,ruled]{algorithm2e}
\usepackage{subcaption}
\usepackage{float}
\usepackage{graphicx}
\usepackage{bm}
\usepackage{times}
\usepackage{balance}
\usepackage{amsmath}
\usepackage{breqn}
\usepackage{cite}
\usepackage{threeparttable}
\usepackage{multirow}
\usepackage{booktabs}
\usepackage{bigstrut}
\usepackage{makecell}
\usepackage{color,soul}
\usepackage{blindtext}
\usepackage{amssymb}

\graphicspath{{figures/}}

\title{\LARGE \bf
    Understanding Spatio-Temporal Relations in Human-Object Interaction using Pyramid Graph Convolutional Network
}

\author{Hao Xing and Darius Burschka
	\thanks{Authors are with Machine Vision and Perception Group, Department of Computer Science,
		Technical University of Munich, Arcisstraße $21$, $80333$ Munich, Germany
		{\tt\small hao.xing@tum.de}, {\tt\small burschka@cs.tum.edu}}%
}%

\begin{document}
	
	\maketitle
	\thispagestyle{empty}
	\pagestyle{empty}

	\begin{abstract}

		Human activities recognition is an important task for an intelligent robot, especially in the field of human-robot collaboration, it requires not only the label of sub-activities but also the temporal structure of the activity. In order to automatically recognize both the label and the temporal structure in sequence of human-object interaction, we propose a novel Pyramid Graph Convolutional Network (PGCN), which employs a pyramidal encoder-decoder architecture consisting of an attention based graph convolution network and a temporal pyramid pooling module for downsampling and upsampling interaction sequence on the temporal axis, respectively. The system represents the 2D or 3D spatial relation of human and objects from the detection results in video data as a graph. To learn the human-object relations, a new attention graph convolutional network is trained to extract  condensed information from the graph representation. To segment action into sub-actions, a novel temporal pyramid pooling module is proposed, which upsamples compressed features back to the original time scale and classifies actions per frame.
		
		We explore various attention layers, namely spatial attention, temporal attention and channel attention, and combine different upsampling decoders to test the performance on action recognition and segmentation. We evaluate our model on two challenging datasets in the field of human-object interaction recognition, i.e. Bimanual Actions and IKEA Assembly datasets. We demonstrate that our classifier significantly improves both framewise action recognition and segmentation, e.g., F1 micro and F1@50 scores on Bimanual Actions dataset are improved by $4.3\%$ and $8.5\%$ respectively.
		

	\end{abstract}

	\linespread{0.9}
    \input{1_Introduction}

    \input{2_Related_work}

\input{3_Methods}

    \input{4_Experiments}

	\input{5_Conclusion}

	\section*{ACKNOWLEDGMENT}
    \vspace*{-0.5\baselineskip}

	We gratefully acknowledge the funding of the Lighthouse Initiative Geriatronics by StMWi Bayern (Project X, grant no. 5140951) and LongLeif GaPa GmbH (Project Y, grant no. 5140953), and our special thanks go to Xiongfei Ma for his help.
	
    \vspace*{-0.5\baselineskip}

	
	
	\bibliographystyle{IEEEtran}
	\bibliography{IEEEexample}

\end{document}

%% file: 1_Introduction.tex
	\section{INTRODUCTION}
    As part of human activities, \textit{human-object interactions} (HOIs) are closely related to the surrounding environment and the objects in the scene. Recognizing HOI in videos is a fundamental task in understanding human activities, in which the sub-activities are segmented and recognized per frame by analyzing the interactive relations between human and objects \cite{morais2021learning}. When human and objects are simply represented by skeleton and center points, these relations naturally form a relation graph in both spatial and temporal dimensions, which can describe their relative positions and dynamic interactions during the activity. Benefiting from the development of deep learning in the field of vision, we can easily build a spatial relation graph by detecting human and object in scenes. However, it is still challenging to discover the temporal structure of sub-actions in a complex task.
    
    Currently, the available graph convolutional networks (GCN) \cite{yan2018spatial, shi2019two} primarily focus on the overall prevalent action being executed, in which only a single action is performed in one set of clips. These methods typically exploit the cascaded structures and can successfully extract and concentrate spatio-temporal features. However, they limit the action recognition task to assigning action labels to the given segments \cite{parsa2020spatio, xing2022skeletal}. Can the extracted spatio-temporal information be used for exploring the temporal structure of activities, i.e., action segmentation?
    
    Regarding this question, we find that it is similar to the difference between image classification and segmentation, where image classification usually adopts a cascade structure, extracts high-level features globally and classify the whole image \cite{krizhevsky2012imagenet}, and image segmentation focuses on the distinction between pixels by upsampling the cascaded features back to the original scale \cite{wu2019fastfcn}.
    
    Hence, in this paper, we propose Pyramid Graph Convolution Network (PGCN) to improve HOI recognition and segmentation by combining the cascaded graph convolutional network with a novel temporal upsampling module, namely temporal pyramid pooling (TPP). Due to the dynamic interactive relations between human and objects, we introduce a novel spatial attention mechanism in GCN to adaptively generate new edges between strongly correlated vertices throughout the activity. 
    The framewise recognition and segmentation capabilities of PGCN are demonstrated with superior quantitative and qualitative performance on two challenging human-object interaction datasets.
    
    Overall, the technical contributions of the paper are:
	\begin{itemize}
		\item We propose Pyramid Graph Convolution Network that utilizes a novel temporal pyramid pooling module to extend the capabilities of GCNs for action segmentation.
		\item We present a new spatial attention mechanism that can improve action recognition by adaptively generating spatial relation graph in dynamic human-objects interaction scenes.
		\item We examine our model on two challenging HOI datasets, the Bimanual Actions and IKEA Assembly datasets. Compared to other good action recognition and segmentation approaches, our model achieves the best quantitative and qualitative performance on both datasets.
	\end{itemize}
	
The rest of the paper is organized as follows: in section \ref{sec:2}, we briefly review existing approaches of GCNs, action segmentation and HOI recognition. Section \ref{sec:3} introduces the proposed PGCN. Section \ref{sec:4} reports experimental results and discussions. Section \ref{sec:5} concludes the paper.

	
	

%% file: 2_Related_work.tex
\section{RELATED WORK}
	\label{sec:2}
\subsection{Graph convolution networks}
Recently, Graph Convolution Networks (GCNs) designed for representation of structured data raise the attention. The GCNs can be categorized into two classes: spatial and spectral. The spatial GCNs operate the graph convolutional kernels directly on spatial graph nodes and their neighborhoods \cite{li2019actional}. Yan et al.~\cite{yan2018spatial} proposed a Spatial-Temporal Graph Convolutional Network (ST-GCN), which extracts spatial feature from the skeleton joints and their naturally connected neighbors and temporal feature from the same joints in consecutive frames. Shi et al.~\cite{shi2019two} introduced a two stream Graph Convolutional Network (2s-GCN) based on ST-GCN, which not only extracts features from skeleton joints but also considers the direction of each joint pair (bone information). Chen et al.~\cite{chen2021channel} proposed a Channel-wise Topology Refinement Graph Convolution  Network (CTR-GCN) that refines a spatial attention mechanism on channel dimension to efficiently learn dynamical features in different channels. 

The spectral GCNs consider the graph convolution in form of spectral analysis \cite{li2015gated}. 
Henaff et al.~\cite{henaff2015deep} developed a spectral network incorporating with graph neural network for the general classification task. Kipf and Welling \cite{kipf2016semi} extends the spectral convolutional network further in the field of semi-supervised learning on graph structured data.

This work follows the spatial GCNs that operate nodes and edges on spatial domain.

\subsection{Action segmentation}
Action Segmentation aims to segment activity by exploring the temporal structure \cite{shou2016}. As part of earlier works, the hidden Markov model (HMM) is often used to find activity temporal structure. 
Pantic et al.\cite{HMM2006} introduced a facial profile recognition scheme combining with HMM to segment facial actions. 
Some other approaches \cite{zelnik2006statistical, xing2021robust} segment action using a sliding window and comparing the similarity between multiple temporal scales. More recently, convolutional neural networks (CNNS) and recurrent neural networks (RNNs) were main streams for action segmentation. For instance, Shou et al. \cite{shou2016} proposed a multi-stage CNN model to classify and localize sub-actions in untrimmed long sequence. Fathi et al. \cite{fathi2013} segment human activities by identifying state changes of objects and materials in the environment using a RNN model. Motivated by the success of temporal convolution in Nature Language Process (NLP) area, many works applied various temporal convolution networks for action segmentation task, such as dilated temporal convolution \cite{hussein2019timeception}, encoder-decoder temporal convolution \cite{Lea2017}. Very recently, attention mechanism from transformer has been successfully applied to action segmentation \cite{zheng2021rethinking}, due to its strong ability of extracting global information. However, the attention mechanism requires known number of involved objects and subjects to define the size of adjacent matrix.

In this work, we take advantage of the attention mechanism to adaptively extract human-objects spatial relations with single subject and known object number and classes.

\subsection{Human-object interaction recognition}
Different from action segmentation task, the HOI recognition task aims at detecting HOI label for whole trimmed action clip. Feichtenhofer et al. \cite{feichtenhofer2016convolutional} introduced a two-stream 2D CNN that utilizes features from both appearance in still images and stacks of optical flow. In a more recent work \cite{carreira2017quo}, authors proposed a two-stream inflated 3D CNN (I3D) that improves the ability of 2D CNNs in extracting spatial-temporal features.
Dreher et al. \cite{dreher2019learning} presented a graph network that uses three  multilayer perceptron (MLP) blocks to update nodes, edges and aggregation features from graph representation of HOI. Authors also published their HOI dataset, namely Bimanual Actions dataset. Asynchronous-Sparse Interaction Graph Networks (ASSIGN) \cite{morais2021learning} is a recent attempt on the HOI recognition task. It used a recurrent graph network that automatically detect the structure of interaction events associated with entities of a sequence of interaction, which are defined as human and objects in a scene. However, the short-term memory of recurrent networks limits their performance in analyzing global temporal structures. In order to expand the receptive field, we adopt dilated convolution layers \cite{wang2016temporal} in the head of our temporal pyramid pooling module, which constrains the implementation in real-time scenarios as it requires relations from future.

%% file: 3_Methods.tex
\section{Pyramid Graph Convolutional Network}
\label{sec:3}
The idea of pyramid graph convolutional network is inspired by upsampling methods for solving image semantic segmentation tasks. A common purpose of both image segmentation and action segmentation is to predict every single elemental unit of the input data by extracting different levels of semantic features and corresponding such features back to the input data to build a segment map. The basic idea of PGCN is to downsample the large-scale data to distill helpful spatial information, which is normally with a smaller temporal scale, and then upsample the distilled information back to the same temporal scale as the input. This is also known as an encoder-decoder structure. 


\subsection{Graph construction}

\begin{figure}[t]
    \centering
    \begin{tabular}{@{}cc@{}}
    \includegraphics[width=0.27\textwidth]{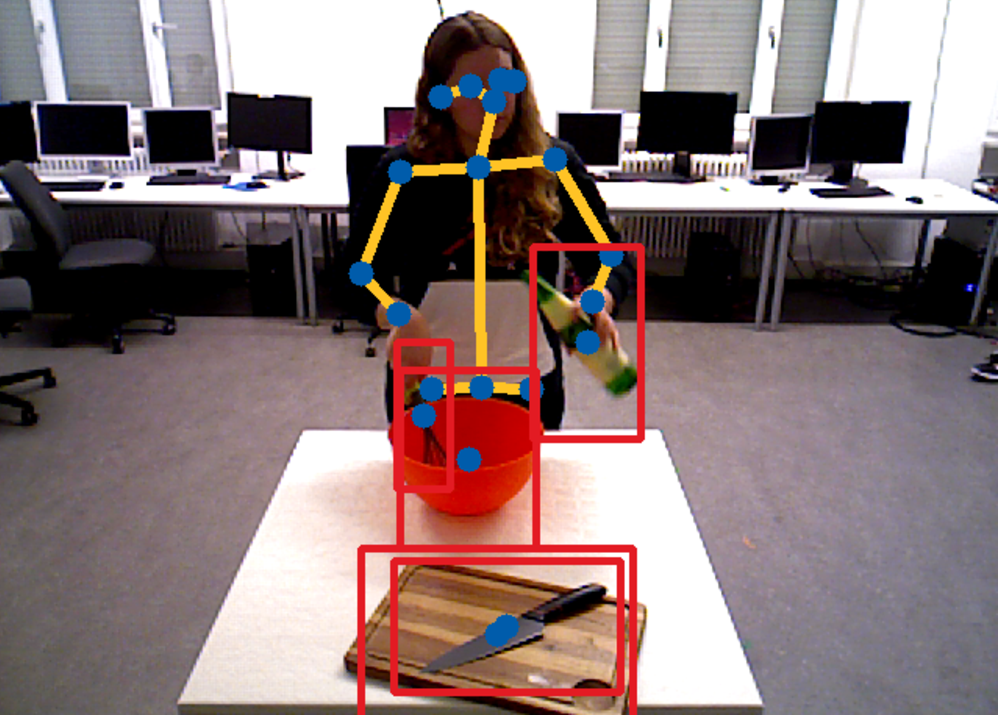} &
    \includegraphics[width=0.2\textwidth]{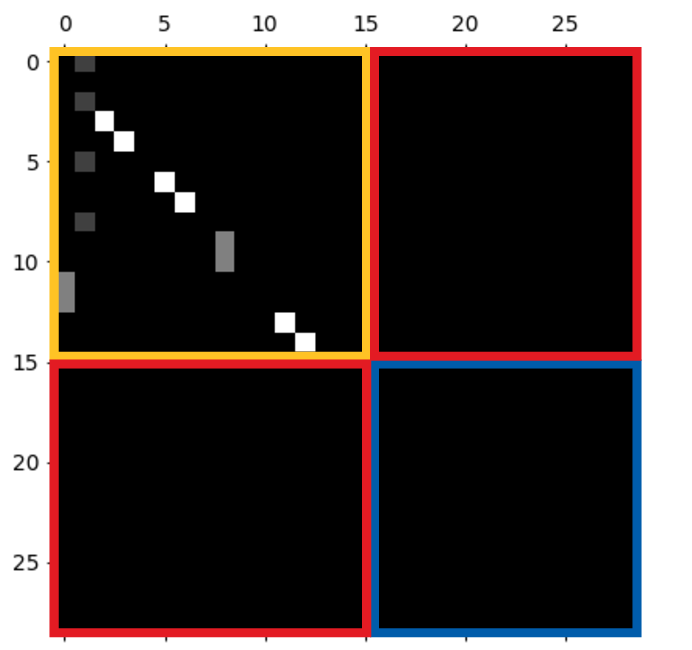} \\
    (a) & (b)
    \end{tabular}
  \caption{\label{fig:graph} The initial spatial relation graph: (a) Spatial graph with notes (blue) and edges (orange) on example of Bimanual Actions dataset \cite{dreher2019learning}; (b) Initial inwards adjacent matrix with skeleton inward edges (orange block), empty human-objects (red blocks) and objects-objects edges (blue block).}
  \label{fig:scene_graph}
  \vspace*{-1.5\baselineskip}
\end{figure}

In order to accurately describe the relationship between people and objects without being affected by texture information, we represent people as skeletons and objects as center points. 
All skeleton joints and object points are vertexes and their connections are represented as edges. Each vertex has inward, outward and self-connecting edges \cite{shi2019two}. The connections between skeleton joints are naturally defined by the pose architecture, with inward connections from each joint to adjacent joints that are closer to the center of the body, and outward connections in reverse. However, object-related connections (human-objects and objects-objects) are challenging due to the dynamic nature of the scene. In this work, we consider that there is no initial connections between objects-objects and between human-objects joints, see Fig.~\ref{fig:scene_graph} (a). 
All edges form a binary adjacency matrix A, in which $a_{ij} = 1$ means vertexes $v_i$ and $v_j$ are connected from $i$ to $j$. Since the initial connections between objects-related pairs of vertexes are not considered, both inward and outward edges are empty, as shown in the Fig~\ref{fig:scene_graph} (b).

Given the adjacency matrix, a spatial scene graph feature map can be obtained by the following equation:
\begin{equation}
    \mathbf{G} = \mathbf{\hat{A}}\cdot\mathbf{F}_{in}
\end{equation}
where $\mathbf{F}_{in}$ is the input skeleton-objects feature map, $\mathbf{G}$ is the graph feature map, and $\mathbf{\hat{A}}$ is column-wise normalization of $\mathbf{A}$. 

\subsection{Attention based graph convolutional encoder}

\begin{figure}[t]
    \centering
    \includegraphics[width=0.32\textwidth]{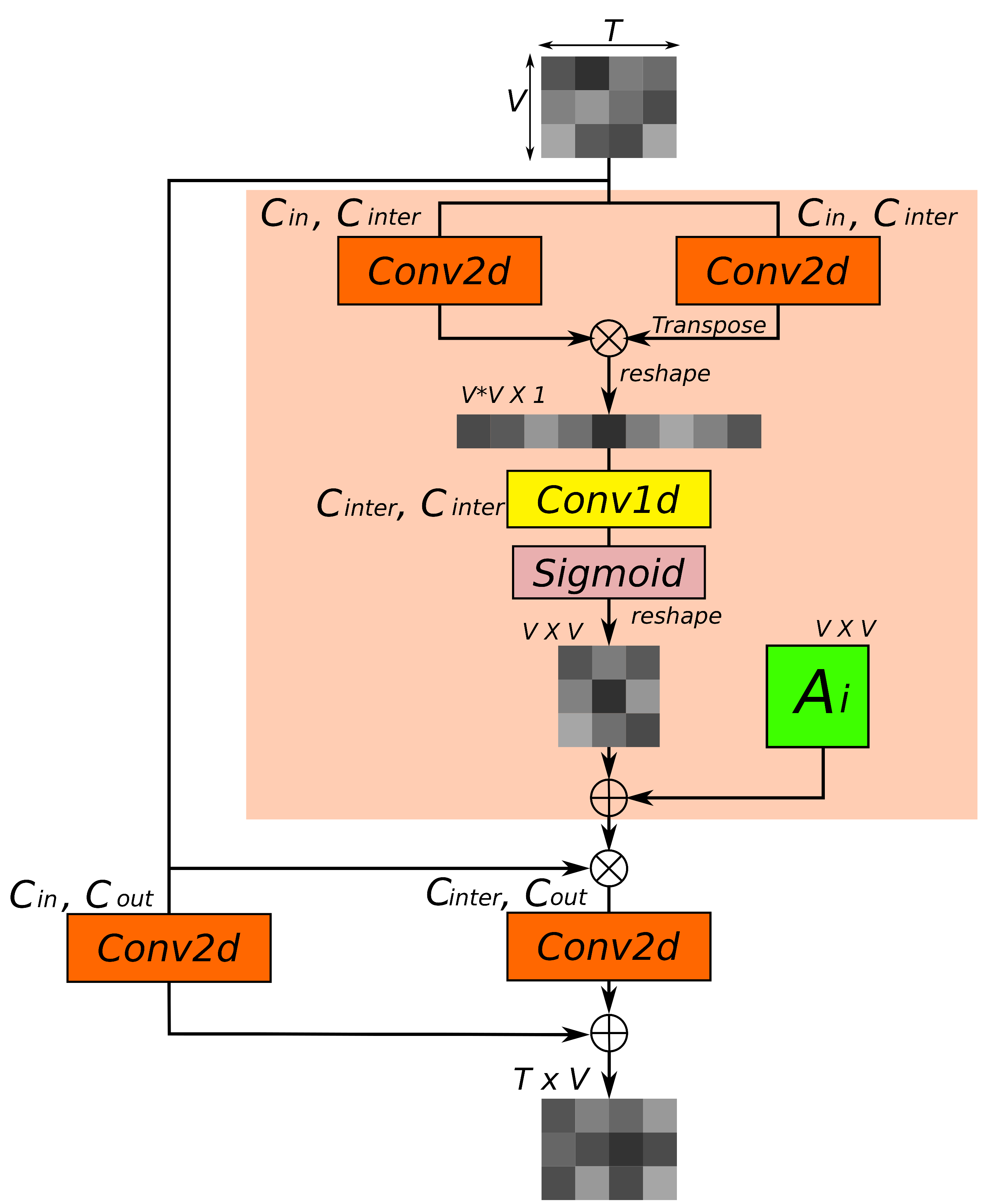}
    \caption{Illustration of the attention unit (orange region) in a spatial convolutional layer}
    \label{fig:spatial_attention}
    \vspace*{-1.5\baselineskip}
\end{figure}

Since the important human-objects interaction information is still missing in the constructed graph, we propose an attention based graph network, which adaptively update the initial adjacent matrix through the attention score map. The attention score is calculated by the dot product between nodes as follows:
\begin{equation}
    M_{ij} = \frac{f_i \cdot f_j^T}{\sqrt{n}}
\end{equation}
where $M$ is the attention mask map, $f$ is the node feature vector and $i, j$ are the indices of nodes. In this work, we find that feeding mask maps into a 1-dimensional convolution layer contributes to the relationship learning process. As shown in Fig. \ref{fig:spatial_attention}, the input feature map is fed into two 2D convolution layers in parallel to generate two output maps with the same size. Their dot product is then fed into a 1D convolution layer with a \emph{sigmoid} activation function to extract the attention mask. The final attention map is generated by the combination of the attention mask with the adjacent matrix as follows:
\begin{equation}
    \mathbf{A}_{final,i} = \mathbf{M}_i + \hat{\mathbf{A}}_i = \mathbf{W}_i(\mathbf{F}_{1,i}^T\cdot \mathbf{F}_{2,i})+\hat{\mathbf{A}}_i
\end{equation}
where $\mathbf{M}$ is the attention mask that is extracted by the 1D convolution kernel on the dot product of feature maps $\mathbf{F}_1$ and $\mathbf{F}_2$, $\mathbf{W}$ is the kernel weight, $\mathbf{A}$ is the adjacent matrices and $i$ is the index of the three connection types (\textit{inwards}, \textit{outwards}, \textit{self-connecting}). In order to give more flexibility to the spatial graph, we set adjacency matrices as learnable parameters with given initial values. 

The output feature map of the spatial attention layer is extended to $C_{out}$ output channels through an additional 2D convolution layer and is merged with the residual stream. The process can be mathematically expressed as follows:
\begin{equation}
    \mathbf{G}_i = \text{Conv2d}(\mathbf{A}_{final,i}\cdot \mathbf{F}_{in}) + \text{res}(\mathbf{F}_{in})
\end{equation}
where $\mathbf{F}_{in}$ is the input feature map, $res$ is the residual layer, and $\mathbf{G}$ is the $i-th$ output graph feature map. The final block output feature map is obtained by summing the outputs of all three types of connections as $\mathbf{G} = \sum_{i=1}^{3}\mathbf{G}_i$

In temporal dimension, we follow the ST-GCN~\cite{yan2018spatial}, i.e., operating a 2D convolution kernel with size $K_t \times 1$ on $C \times T \times V$ feature maps, where $K_t$ is set to $9$ in this work.

Given the defined spatial, temporal layer, an attention based graph convolutional block is formed. 
In the encoder, we employ 10 basic blocks and connect them through the usual cascade structure, as introduced in \cite{yan2018spatial, shi2019two}. 

\subsection{Temporal pyramid upsampling decoder}


\begin{figure*}[ht]
\centering
\includegraphics[width=0.82\textwidth]{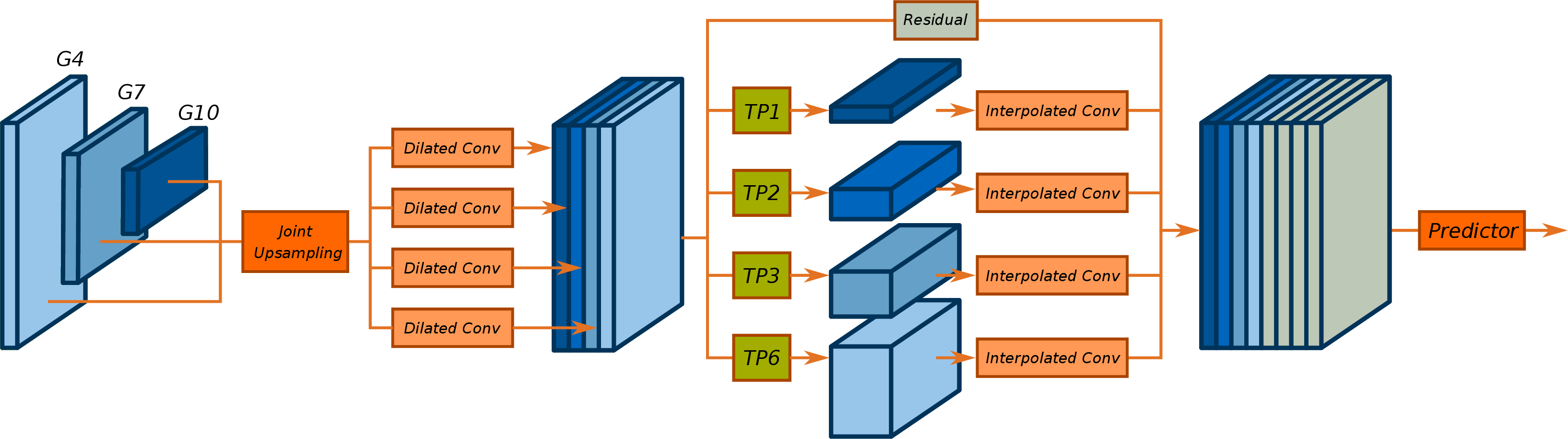}
\caption{Framework of temporal pyramid pooling decoder with three input graph feature maps: $\mathbf{G}^4$, $\mathbf{G}^7$ and $\mathbf{G}^{10}$, where $TP\ i$ is temporal pooling block with output size $i$.}
\label{fig:framework}
\vspace*{-1.5\baselineskip}
\end{figure*}

Given the introduced encoder, three graph feature maps $\mathbf{G}_{in}=\{\mathbf{G}^4, \mathbf{G}^7, \mathbf{G}^{10} \}$ of $4$-th, $7$-th and $10$-th blocks are jointly taken as input into the temporal upsampling module, which contain different level semantic information. Since these feature maps have different size, we unify the number of channels through a 2D convolution kernel and interpolate all feature maps to the initial time scale concatenate them along channel dimension. A segmentation feature extraction is performed through four parallel dilated convolution operations \cite{wang2016temporal} as following:
\begin{equation}
    \mathbf{G}_{out} = \sqcup_{i=1}^{4}\sigma(\mathbf{G}_{in, u}\mathbf{W}^d_i + \mathbf{B}^d_i)
\end{equation}
where $\mathbf{G}_{out}$ is output graph feature map, $\mathbf{G}_{in, u}$ is upscaled input graph feature map, $\sqcup_{i=1}^{4}$ indicates the concatenation operation with $4$ streams, $\sigma$ is the \textit{ReLU} activation function, $\mathbf{W}^d_i$ and $\mathbf{B}^d_i$ are parameters of $i$-th dilated convolutional kernel.

In order to extract a global contextual prior for prediction, a temporal pyramid pooling module is utilized before the predictor. 
In semantic segmentation tasks regarding images, global average pooling is a general choice as the global contextual prior. 
However, in the action segmentation, the features in temporal and spatial dimensions need to be handled differently. 
Since the final segmentation is in temporal dimension, i.e., the sequence of predicted labels per frame, four pyramid temporal average pooling blocks of different scales are first performed along the temporal dimension to extract a segment prior with multiple receptive fields. 

Given the time series dilated graph feature map $\mathbf{G}_{out}\in\mathbb{R} ^{N\times T}$ with $N$ spacial nodes and $T$ frames, it can be represented as a set of time segments at level $i$ as $\mathbf{G}_{out} = \{  \mathbf{G}_1, \cdots, \mathbf{G}_i \}$. A temporal filter with average pooling operator is applied to each time segment $[t_{min}, t_{max}]$ and provides a single feature vector for each segment as:
\begin{equation}
    \mathcal{O}(\mathbf{G}_i) = \frac{\sum_{t_{min}}^{t_{max}}g_t^i}{t_{max}-t_{min}}
\end{equation}

Then, a convolution layer is performed in spatial dimension to extract global spatial information with various temporal scales as following:
\begin{equation}
    \mathbf{F}_{out} = \sigma(\mathbf{G}_{out}\mathbf{W}_s + \mathbf{B}_s)
\end{equation}
where $\mathbf{W}_s\in\mathbb{R}^{k\times 1}$ and $\mathbf{B}_s\in\mathbb{R}^{k\times 1}$ are parameters of the spatial convolutional kernel, and $k\times 1$ indicates the kernel size.

The four low-dimension output feature maps are directly upsampled by bilinear interpolation to have the same temporal and spatial lengths as the original feature maps. At last, four different levels of features are concatenated with the residual feature map. After obtaining the feature map containing global contextual prior with various scales and framewise local features, a convolution based predictor is used to generate framewise interaction labels. The framework is illustrated in Fig.~\ref{fig:framework}.

%% file: 4_Experiments.tex
\section{EXPERIMENTS AND RESULTS}
\label{sec:4}

To evaluate the performance of proposed model, we experiment on two challenging human-object interaction recognition datasets: Bimanual Actions dataset~\cite{dreher2019learning} and IKEA Assembly dataset~\cite{ikea2020}.
We first perform detailed ablation study on the Bimanual Actions dataset \cite{dreher2019learning} to examine the contributions of the proposed model components. Then, we evaluate the final model on both datasets and compare the results with other state-of-the-art methods.

\subsection{Dataset}
\textbf{Bimanual Actions Dataset} \cite{dreher2019learning} was build for human object interaction detection in third-person perspective. It contains 540 recordings with a total runtime of 2 hours 18 minutes. It has framewise predictions of 12 objects (3D bounding boxes) and 6 subjects (3D skeletons), including the both hands of subjects as one of 14 possible interaction categories. In each record, a single person is performing a complex daily task in one of the two set environments, namely kitchen and workshop. The authors of the dataset recommend a benchmark: \textbf{leave-one-subject-out} cross-validation that contains records from one subject for validation and the rest subjects for training.

\textbf{IKEA Assembly Dataset} \cite{ikea2020} is a more challenging and complex human-object interaction dataset, which contains a total of 16,764 annotated actions   with an average of 150 frames per action ($\sim$ $35.27$h). The authors proposed a \textbf{cross-environment} benchmark, in which the test environments do not appear in the trainset and vise-versa. 
The trainset and testset consist of 254 and 117 scans, respectively.

\begin{table*}[t!]
  \caption{The F1 score of framewise prediction and F1@k score of action segmentation using original baseline model and models with different modifications in each unit}
  \label{tab:ablation_study}
  \centering
  \begin{tabular}{c c c | c c | c c | c c c}
    \toprule
    \multicolumn{3}{c|}{Encoder $^a$} & \multicolumn{2}{c|}{Decoder $^b$} &
    \multicolumn{5}{c}{Evaluation Metrics $^c$}
    \\         \midrule

    Spatial & Temporal & Channel & TPP & Fast-FCN & F1 macro ($\%$) & F1 micro ($\%$) & F1@10 ($\%$) & F1@25 ($\%$) & F1@50 ($\%$)\\
    \midrule
      - & - & - & - & \checkmark & \underline{65.28} & 80.01 & 66.51 & 62.44 & 51.37  \\
      - & - & - & \checkmark & - & 65.17 & \underline{81.80} & \underline{86.36} & \underline{83.66} & \underline{71.84}  \\
      \checkmark & - & - & - & \checkmark & 70.92 & 83.09 & 70.38 & 66.26 & 66.27  \\
      \checkmark & - & - & \checkmark & - &\underline{\textbf{81.50}} & \underline{\textbf{86.92}} & \underline{\textbf{88.38}} & \underline{\textbf{85.06}} & \underline{73.88}  \\
      - & \checkmark & - & - & \checkmark & 75.93 & 83.42 & 67.83 & 63.51 & 52.68 \\
      - & \checkmark & - & \checkmark & - & \underline{77.26} & \underline{84.94} & \underline{78.77} & \underline{75.46} & \underline{61.43}\\
      - & - & \checkmark & - & \checkmark & \underline{70.16} & \underline{83.56} & 74.16 & 70.55 & 58.92\\
      - & - & \checkmark & \checkmark & - & 67.39 & 82.50  & \underline{88.23} & \underline{84.86} & \underline{\textbf{74.10}}\\
      \checkmark & \checkmark & - & \checkmark  & - & 80.29 & 85.25 & 84.38 & 81.46 & 68.55 \\
      \checkmark & - & \checkmark & \checkmark  & - & 69.65 & 80.57 & 84.09 & 81.07 & 66.99  \\
      - & \checkmark & \checkmark & \checkmark  & - & 71.42 & 82.75 & 85.36 & 81.45 & 69.21 \\
      \checkmark & \checkmark & \checkmark & \checkmark & - & 72.39 & 83.63 & 85.68 & 81.94 & 70.86 \\
    \bottomrule
  \end{tabular}
  \scriptsize
  \begin{tablenotes}
    \item[a]$^a$ We compare the performance of attention layer in the encoder setup on different dimensions, namely spatial, temporal and channel.
    \item[a]$^b$ The decoder is the common Fast-FCN \cite{wu2019fastfcn} when there is no temporal pyramid pooling block.
    \item[a]$^c$ The best results comparing all modifications are in \textbf{bold}; The best results between TPP and Fast-FCN in the decoder setup are \underline{underlined}
  \end{tablenotes}
\vspace*{-1.5\baselineskip}
\end{table*}

\subsection{Experimental settings}
We evaluate the proposed model on the task: HOI framewise recognition and temporal segmentation. Two main evaluation metrics, i.e., F1-score and F1@k, are selected for framewise recognition and segmentation, respectively. the F1-score is formulated as: $\text{F1} = tp/(tp+0.5(fp+fn))$ where $tp$ means true positive predictions, $fp$ and $fn$ refer to false positive and false negative predictions, respectively.
For the F1@k score, common values of $k = 0.10$, $0.25$, and $0.50$ are used, in which the true or false positive for each predicted segment is determined by comparing the intersection over union (IoU) with threshold $\tau =k/100$. Incorrect predictions and missed ground-truth segments are counted as false positive and false negative, respectively. Moreover, for the multi-class prediction task, micro-average and macro-average over F1-scores of all classes are adopted as the framewise recognition metric. In the experiment comparing popular methods on IKEA Assembly dataset, we used top1 and macro-recall metrics to evaluate models.

For the Bimanual Action dataset \cite{dreher2019learning}, we use the center of 3D object bounding boxes and 3D human skeleton data released by authors \cite{dreher2019learning}, leave subject $1$ out for validation in ablation study, and do leave-one-subject-out cross-validation in comparison with other popular methods. For the IKEA Assembly dataset \cite{ikea2020}, we use the offered center of 2D object bounding boxes and 2D human skeleton data, and follow the cross-environment benchmark.

The models and experiments are implemented and conducted on the PyTorch deep learning framework with a single NVIDIA-2070 GPU. The optimization strategy is selected to be the widely used stochastic gradient descent (SGD) with Nesterov momentum ($0.9$). Cross-entropy is applied as the loss function for the gradient back propagation. $32$ batch size is applied for both training and testing. The weight decay is set to be 0.0001. The training process contains 60 epochs in total. The initial learning rate is set to be $0.1$, and is divided by 10 at the $20$-th and $40$-th epoch.


\subsection{Ablation studies}

We examine the contribution of proposed components to the framewise HOI recognition and segmentation with leaving subject $1$ on the Bimanual Actions dataset \cite{dreher2019learning}. The baseline is the single joint stream of 2s-AGCN \cite{shi2019two}. 
 
In order to get the best performance of the attention unit, we evaluate the unit on the dimension of spatial, temporal, channel and their combinations. The proposed temporal pyramid pooling module is compared with the baseline FastFCN~\cite{wu2019fastfcn}.

The results of ablation studies are shown in Table~\ref{tab:ablation_study}, where the performance of each setting is quantified by F1 and F1@k scores. From the F1@k score in the right column of the table, it is obviously that the temporal pyramid pooling block shows improvement on relation segmentation for each specific setting. Hence, we include the temporal pyramid pooling block for rest experiments on combined attention layers. 

From the F1 scores in the middle column, we can see that all proposed components improve the performance of the baseline model AGCN~\cite{shi2019two} on framewise recognition. Moreover, since the spatial attention unit extracts the basic features representing spatial distribution and relations of nodes per frame, model with spatial attention unit has the best performance among all model settings. The temporal attention unit extracts temporal relations between consecutive frames, which is beneficial for segmentation-known action recognition rather than the action segmentation. The channel attention layer focuses on the importance of distinguishing between channels, which is conducive to the classification of an entire clip of single action rather than the segmentation. The performance of combined models suffers from poor attention layers, namely temporal and channel attention layers.

\begin{figure}[t]
  \centering
  \includegraphics[width=0.9\linewidth]{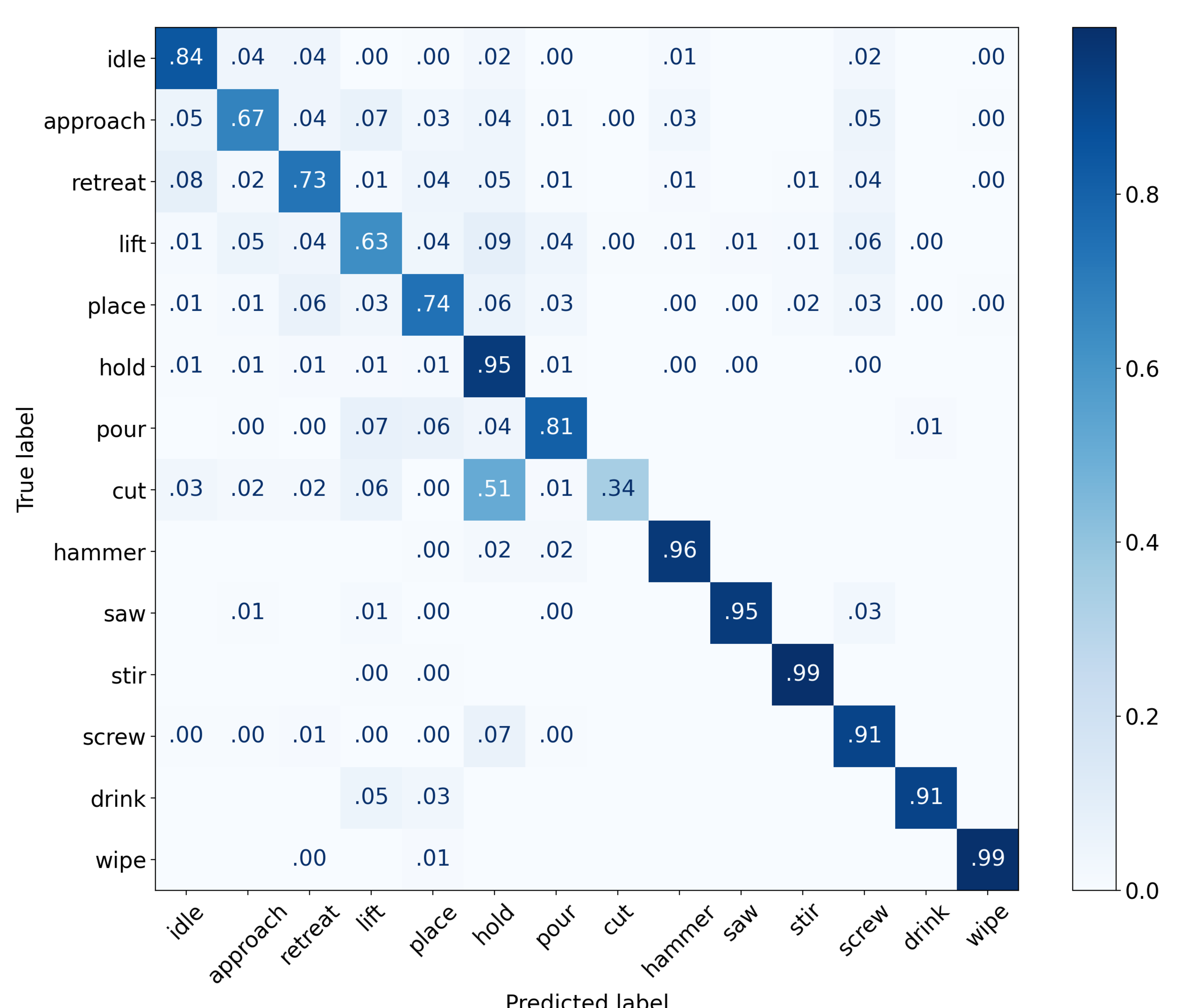}
  \caption{Normalized confusion matrix for the top prediction of accumulative framewise classification correctness over all folds on Bimanual Actions dataset \cite{dreher2019learning}. }\label{fig:confusion}
   \vspace*{-1.5\baselineskip}
\end{figure}

Besides F1 scores, we also evaluate the Top-1 accuracy of the proposed model on Bimanual Actions dataset \cite{dreher2019learning}. Fig.~\ref{fig:confusion} depicts the normalized confusion matrices for the top prediction. A major confusion of the classifier is the prediction of \textit{hold} while the true action is \textit{cut}. The cause for the wrong prediction is that we use \textit{wrist} joints to represent \textit{hands}, which has 
a small range of motion and is easily mistaken for \textit{holding} a \textit{knife}. Therefore, the prediction of \textit{cut} usually has a large range of motion and is rarely falsely recognized from the action of \textit{hold}. There is also a number of confusion between actions \textit{approach}, \textit{retreat} \textit{lift} and \textit{place}. This is suffering from an unstable object bounding box, namely the object detection method. Additionally, \textit{approach} and \textit{retreat} are usually executed fast (sometimes within $5$ frames). An example can be found in qualitative results. These problems can be mitigated by stable object detection and pose estimation methods. However, it is not the main focus of this work and will not be addressed.

\begin{figure*}[t]
  \centering
  \includegraphics[width=0.8\textwidth]{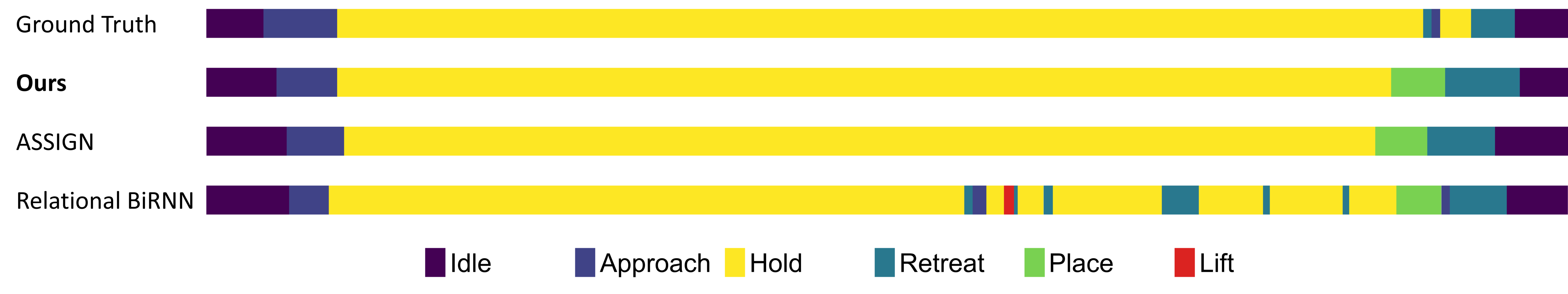}
  \caption{Comparison of the qualitative results on Bimanual Actions dataset \cite{dreher2019learning} for a \textit{sawing} example}\label{fig:seg_view_bimacs}
\end{figure*}

\begin{figure*}[t]
  \centering
  \includegraphics[width=0.8\textwidth]{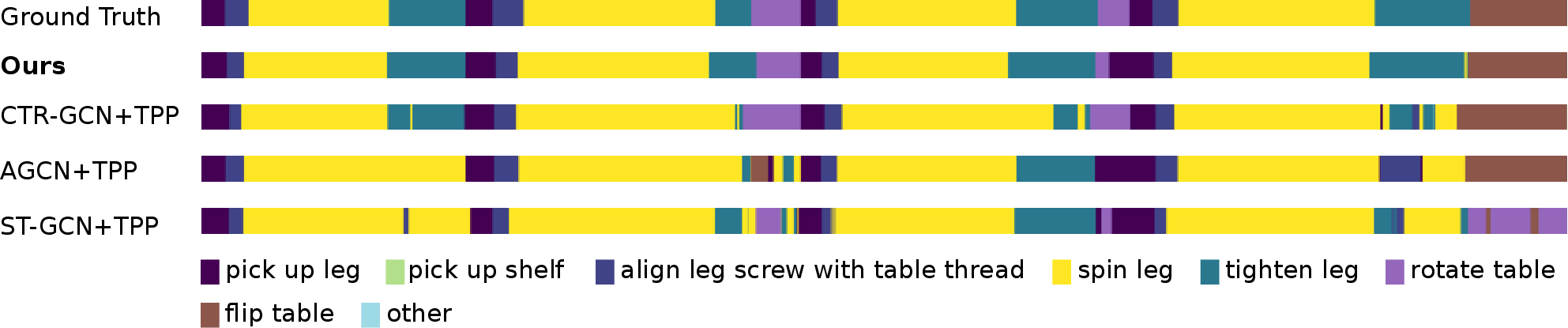}
  \caption{Comparison of the qualitative results on IKEA Assembly dataset \cite{ikea2020} for an \textit{assembly side table} example}\label{fig:seg_view_IKEA}
\vspace*{-1.5\baselineskip}

\end{figure*}

\subsection{Comparison with the state-of-the-art}

The proposed PGCN model is compared with the state-of-the-art action recognition and segmentation methods on Bimanual Actions \cite{dreher2019learning} and IKEA Assembly datasets \cite{ikea2020}. 
The methods used for comparison include the model proposed by Dreher et al.~\cite{dreher2019learning}, Independent BiRNN, Relational BiRNN, ASSIGN~\cite{morais2021learning} and several popular graph convolutional networks: ST-GCN~\cite{yan2018spatial}, AGCN~\cite{shi2019two} and CTR-GCN~\cite{chen2021channel} combining with two decoders, namely FastFCN \cite{wu2019fastfcn} and the proposed temporal pyramid pooling (TPP) module.


\begin{table}[t]
  \caption{Comparison of framewise action recognition with state-of-the-art methods on Bimanual Actions dataset \cite{dreher2019learning}}\label{tab:sota_framewise}
  \centering
  \begin{tabular}{c c c}
    \toprule
      Model & F1 macro (\%) & F1 micro (\%) \\
    \midrule
      Dreher et al. \cite{dreher2019learning} & $63.0$ & $64.0$ \\
      AGCN+FastFCN & $65.3$ & $80.0$ \\
      AGCN+TPP & $65.2$ & $81.8$ \\
      ST-GCN+FastFCN & $68.7$ & $82.5$ \\
      ST-GCN+TPP & $69.3$ & $82.7$ \\
      CTR-GCN+FastFCN & $71.1$ & $82.3$ \\
      CTR-GCN+TPP & $72.0$ & $82.9$ \\
      Independent BiRNN \cite{morais2021learning} & $74.8$ & $76.7$ \\
      Relational BiRNN \cite{morais2021learning} & $77.5$ & $80.3$ \\
      ASSIGN \cite{morais2021learning} & ${79.8}$ & $82.6$ \\
    \midrule
      \textbf{PGCN} (Ours) & $\mathbf{81.5}$ & $\mathbf{86.9}$ \\
    \bottomrule
  \end{tabular}
      \vspace*{-1.\baselineskip}
\end{table}

\begin{table}[t]
  \caption{Cross validation results of action segmentation in comparison with state-of-the-art methods on Bimanual Actions dataset \cite{dreher2019learning}}\label{tab:sota_seg}
  \centering
  \begin{tabular}{c c c c}
    \toprule
      Model & F1@10 (\%) & F1@25 (\%) & F1@50 (\%) \\
    \midrule
      Dreher et al. \cite{dreher2019learning} & $40.6 \pm 7.2$ & $34.8 \pm 7.1$ & $22.2 \pm 5.7$ \\
      Independent BiRNN \cite{morais2021learning} & $74.8 \pm 7.0$ & $72.0 \pm 7.0$ & $61.8 \pm 7.3$ \\
      CTR-GCN+FastFCN & $74.9 \pm 8.1$ & $72.2 \pm 8.7$ & $66.6 \pm 11.4$ \\
      Relational BiRNN \cite{morais2021learning} & $77.7 \pm 3.9$ & $75.0 \pm 4.2$ & $64.8 \pm 5.3$ \\
      ASSIGN \cite{morais2021learning} & $84.0 \pm 2.0$ & $81.2 \pm \mathbf{2.0}$ & $68.5 \pm \mathbf{3.3}$ \\
      CTR-GCN+TPP & $84.8 \pm 3.2$ & $82.1 \pm 4.0$ & $73.5 \pm 5.6$ \\
    \midrule
      \textbf{PGCN} (Ours) & $\mathbf{88.5} \pm \mathbf{1.1}$ &$\mathbf{85.5} \pm \mathbf{2.0}$ & $\mathbf{77.0} \pm 3.4$\\
    \bottomrule
  \end{tabular}
\vspace*{-1.5\baselineskip}
\end{table}

The performance in terms of F1 score and F1@k on the Bimanual Actions dataset \cite{dreher2019learning} are listed in Table~\ref{tab:sota_framewise} and Table~\ref{tab:sota_seg}, respectively. The PGCN outperforms both the state-of-the-art and baselines in every configuration of the F1 and F1@k measure, e.g., the F1 macro and micro score are improved by $1.7 \%$ and $4.3 \%$ respectively. Moreover, it can be observed that the proposed temporal pyramid pooling block improves significantly the performance in terms of F1@k score (by $4.5 \%$, $4.3 \%$ and $8.5 \%$ compared to ASSIGN, respectively), which again confirms its efficiency in action segmentation. The ASSIGN uses the Bi-directional Gated Recurrent Unit to combine information from consecutive frames, which limitedly enhances the extraction of temporal information and causes the shift-segmentation. Other methods employ separate segmentation label, which is lack of temporal information and leads to an over-segmentation case. More evidences can be found in the qualitative results.

Table~\ref{tab:sota_seg_IKEA} presents the top-1 accuracy, micro-recall and F1@k score on IKEA Assembly dataset \cite{ikea2020}. In terms of top-1, and all three F1@k scores, PGCN outperforms all other popular methods. This further demonstrates that our modeling entities with spatial attention and temporal pyramid pooling modules is a more competitive way to recognize and segment action per frame. Due to the uneven distribution of the dataset~\cite{ikea2020}, the macro-recall of all methods is low.

\begin{table}[t]
  \caption{Framewise recognition and segmentation results in terms of top-1 accuracy, macro-recall, and F1@k on IKEA Assembly dataset \cite{ikea2020}}\label{tab:sota_seg_IKEA}
  \centering
  \begin{tabular}{c| c c | c c c}
    \toprule
      \multirow{2}{*}{Model} & \multirow{2}{*}{top 1} & \multirow{2}{*}{macro} & \multicolumn{3}{c}{F1@ ($\%$)}  \\
      & &  & 10  & 25 & 50  \\
    \midrule
    HCN~\cite{li2018co} & $39.15$ & $28.18$ & - & - & -  \\
    ST-GCN~\cite{yan2018spatial} &  $43.40$ & $26.54$ & - & - & -  \\
    multiview+HCN~\cite{ikea2020} &  $64.25$ & $\mathbf{46.33}$ & - & - & -  \\
    ST-GCN+TPP & $68.92$ & $25.63$ & $66.92$ & $59.66$ & $41.33$  \\
    AGCN+TPP & $70.53$ & $27.79$ & $76.32$ & $69.85$ & $52.14$  \\
    CTR-GCN+TPP & $78.70$ & $37.98$ & $78.84$ & $72.68$ & ${54.40}$   \\
    \midrule
      \textbf{PGCN} (Ours) & $\mathbf{79.35}$ &${38.29}$ & $\mathbf{81.53}$ & $\mathbf{76.28}$ & $\mathbf{58.07}$\\
    \bottomrule
  \end{tabular}
 \vspace*{-1.5\baselineskip}
\end{table}

\subsection{Qualitative results}
We present the detail outputs of PGCN model and related methods on examples from Bimanual Actions \cite{dreher2019learning} and IKEA Assembly dataset \cite{ikea2020}. Fig.~\ref{fig:seg_view_bimacs} shows a simple example of \textit{sawing} in Bimanual Actions \cite{dreher2019learning}, where both PGCN and ASSIGN have a stronger ability to prevent over-segmentation than Relational BiRNN. Our PGCN demonstrates more accurate segmentation than ASSIGN, which even recognizes correctly the frame index between \textit{Approach} and \textit{Hold} ($0$ frame error) in the example. As aforementioned, brief movement can easily lead to false predictions, see the end of the ground-truth and predictions.


Besides the simple example, Fig.~\ref{fig:seg_view_IKEA} presents a segmentation example of complex \textit{assembly side table} task on the IKEA Assembly dataset \cite{ikea2020}, where we compare the qualitative performance of methods with the same decoder and different encoders. It can be seen that our PGCN model prevents under-segmentation better than other models with the same decoder, which further demonstrates the effectiveness of our spatial attention unit. From simple to complex tasks, our model demonstrates strong and stable performance.

%% file: 5_Conclusion.tex
\section{CONCLUSIONS}
\label{sec:5}
In this work, we introduce a novel pyramid graph convolutional network for understanding human-object interaction relation sequences via action recognition and segmentation, which includes a spatial attention graph convolutional encoder and a temporal pyramid pooling decoder.

The two components are complementary to each other, i.e., the spatial attention mechanism provides high-level spatial relations between human and objects to the decoder, and the temporal pyramid pooling decoder upsamples these spatial features to the original time-scale and predict framewise labels. Experimental analysis into PGCN's components shows that the new attention layer improves the accuracy of action recognition, further mitigate under-segmentation, and the new temporal pyramid block has strong ability to prevent action over- and shift-segmentation. Results on two HOI datasets with different input formats (2D and 3D) show that PGCN has a general capability that can be implemented on other structural-represented domains. Our future work will explore multi-persons involved HOI recognition and segmentation, and try to overcome the constrains of temporal pyramid pooling model and implement our model in real-time Human-Robot Collaboration tasks.

%% file: main.bbl
\begin{thebibliography}{10}
\providecommand{\url}[1]{#1}
\csname url@rmstyle\endcsname
\providecommand{\newblock}{\relax}
\providecommand{\bibinfo}[2]{#2}
\providecommand\BIBentrySTDinterwordspacing{\spaceskip=0pt\relax}
\providecommand\BIBentryALTinterwordstretchfactor{4}
\providecommand\BIBentryALTinterwordspacing{\spaceskip=\fontdimen2\font plus
\BIBentryALTinterwordstretchfactor\fontdimen3\font minus
  \fontdimen4\font\relax}
\providecommand\BIBforeignlanguage[2]{{%
\expandafter\ifx\csname l@#1\endcsname\relax
\typeout{** WARNING: IEEEtran.bst: No hyphenation pattern has been}%
\typeout{** loaded for the language `#1'. Using the pattern for}%
\typeout{** the default language instead.}%
\else
\language=\csname l@#1\endcsname
\fi
#2}}

\bibitem{morais2021learning}
R.~Morais, V.~Le, S.~Venkatesh, and T.~Tran, ``Learning asynchronous and sparse
  human-object interaction in videos,'' in \emph{Proceedings of the IEEE/CVF
  Conference on Computer Vision and Pattern Recognition}, 2021, pp.
  16\,041--16\,050.

\bibitem{yan2018spatial}
S.~Yan, Y.~Xiong, and D.~Lin, ``Spatial temporal graph convolutional networks
  for skeleton-based action recognition,'' in \emph{Proceedings of the AAAI
  conference on artificial intelligence}, 2018.

\bibitem{shi2019two}
L.~Shi, Y.~Zhang, J.~Cheng, and H.~Lu, ``Two-stream adaptive graph
  convolutional networks for skeleton-based action recognition,'' in
  \emph{Proceedings of the IEEE/CVF Conference on Computer Vision and Pattern
  Recognition}, 2019, pp. 12\,026--12\,035.

\bibitem{parsa2020spatio}
B.~Parsa, B.~Dariush, \emph{et~al.}, ``Spatio-temporal pyramid graph
  convolutions for human action recognition and postural assessment,'' in
  \emph{Proceedings of the IEEE/CVF Winter Conference on Applications of
  Computer Vision}, 2020, pp. 1080--1090.

\bibitem{xing2022skeletal}
H.~Xing and D.~Burschka, ``Skeletal human action recognition using hybrid
  attention based graph convolutional network,'' in \emph{26th International
  Conference on Pattern Recognition (ICPR)}, 2022.

\bibitem{krizhevsky2012imagenet}
A.~Krizhevsky, I.~Sutskever, and G.~E. Hinton, ``Imagenet classification with
  deep convolutional neural networks,'' \emph{Advances in neural information
  processing systems}, vol.~25, 2012.

\bibitem{wu2019fastfcn}
H.~Wu, J.~Zhang, K.~Huang, K.~Liang, and Y.~Yu, ``Fastfcn: Rethinking dilated
  convolution in the backbone for semantic segmentation.'' \emph{CoRR}, vol.
  abs/1903.11816, 2019.

\bibitem{li2019actional}
M.~Li, S.~Chen, X.~Chen, Y.~Zhang, Y.~Wang, and Q.~Tian, ``Actional-structural
  graph convolutional networks for skeleton-based action recognition,'' in
  \emph{Proceedings of the IEEE/CVF Conference on Computer Vision and Pattern
  Recognition}, 2019, pp. 3595--3603.

\bibitem{chen2021channel}
Y.~Chen, Z.~Zhang, C.~Yuan, B.~Li, Y.~Deng, and W.~Hu, ``Channel-wise topology
  refinement graph convolution for skeleton-based action recognition,'' in
  \emph{Proceedings of the IEEE/CVF International Conference on Computer
  Vision}, 2021, pp. 13\,359--13\,368.

\bibitem{li2015gated}
Y.~Li, D.~Tarlow, M.~Brockschmidt, and R.~S. Zemel, ``Gated graph sequence
  neural networks,'' in \emph{4th International Conference on Learning
  Representations, {ICLR} 2016}, 2016.

\bibitem{henaff2015deep}
M.~Henaff, J.~Bruna, and Y.~LeCun, ``Deep convolutional networks on
  graph-structured data,'' \emph{arXiv preprint arXiv:1506.05163}, 2015.

\bibitem{kipf2016semi}
T.~N. Kipf and M.~Welling, ``Semi-supervised classification with graph
  convolutional networks,'' in \emph{5th International Conference on Learning
  Representations, {ICLR} 2017}, 2017.

\bibitem{shou2016}
Z.~Shou, D.~Wang, and S.-F. Chang, ``Temporal action localization in untrimmed
  videos via multi-stage cnns,'' in \emph{2016 IEEE Conference on Computer
  Vision and Pattern Recognition (CVPR)}, 2016, pp. 1049--1058.

\bibitem{HMM2006}
M.~Pantic and I.~Patras, ``Dynamics of facial expression: recognition of facial
  actions and their temporal segments from face profile image sequences,''
  \emph{IEEE Transactions on Systems, Man, and Cybernetics, Part B
  (Cybernetics)}, vol.~36, no.~2, pp. 433--449, 2006.

\bibitem{zelnik2006statistical}
L.~Zelnik-Manor and M.~Irani, ``Statistical analysis of dynamic actions,''
  \emph{IEEE Transactions on Pattern Analysis and Machine Intelligence},
  vol.~28, no.~9, pp. 1530--1535, 2006.

\bibitem{xing2021robust}
H.~Xing, Y.~Xue, M.~Zhou, and D.~Burschka, ``Robust event detection based on
  spatio-temporal latent action unit using skeletal information,'' in
  \emph{2021 IEEE/RSJ International Conference on Intelligent Robots and
  Systems (IROS)}.\hskip 1em plus 0.5em minus 0.4em\relax IEEE, 2021, pp.
  2941--2948.

\bibitem{fathi2013}
A.~Fathi and J.~M. Rehg, ``Modeling actions through state changes,'' in
  \emph{2013 IEEE Conference on Computer Vision and Pattern Recognition}, 2013,
  pp. 2579--2586.

\bibitem{hussein2019timeception}
N.~Hussein, E.~Gavves, and A.~W. Smeulders, ``Timeception for complex action
  recognition,'' in \emph{Proceedings of the IEEE/CVF Conference on Computer
  Vision and Pattern Recognition}, 2019, pp. 254--263.

\bibitem{Lea2017}
C.~Lea, M.~D. Flynn, R.~Vidal, A.~Reiter, and G.~D. Hager, ``Temporal
  convolutional networks for action segmentation and detection,'' in \emph{2017
  IEEE Conference on Computer Vision and Pattern Recognition (CVPR)}, 2017, pp.
  1003--1012.

\bibitem{zheng2021rethinking}
S.~Zheng, J.~Lu, H.~Zhao, X.~Zhu, Z.~Luo, Y.~Wang, Y.~Fu, J.~Feng, T.~Xiang,
  P.~H. Torr, \emph{et~al.}, ``Rethinking semantic segmentation from a
  sequence-to-sequence perspective with transformers,'' in \emph{Proceedings of
  the IEEE/CVF conference on computer vision and pattern recognition}, 2021,
  pp. 6881--6890.

\bibitem{feichtenhofer2016convolutional}
C.~Feichtenhofer, A.~Pinz, and A.~Zisserman, ``Convolutional two-stream network
  fusion for video action recognition,'' in \emph{Proceedings of the IEEE
  conference on computer vision and pattern recognition}, 2016, pp. 1933--1941.

\bibitem{carreira2017quo}
J.~Carreira and A.~Zisserman, ``Quo vadis, action recognition? a new model and
  the kinetics dataset,'' in \emph{proceedings of the IEEE Conference on
  Computer Vision and Pattern Recognition}, 2017, pp. 6299--6308.

\bibitem{dreher2019learning}
C.~R. Dreher, M.~W{\"a}chter, and T.~Asfour, ``Learning object-action relations
  from bimanual human demonstration using graph networks,'' \emph{IEEE Robotics
  and Automation Letters}, vol.~5, no.~1, pp. 187--194, 2019.

\bibitem{wang2016temporal}
P.~Wang, Y.~Cao, C.~Shen, L.~Liu, and H.~T. Shen, ``Temporal pyramid
  pooling-based convolutional neural network for action recognition,''
  \emph{IEEE Transactions on Circuits and Systems for Video Technology},
  vol.~27, no.~12, pp. 2613--2622, 2016.

\bibitem{ikea2020}
Y.~Ben-Shabat, X.~Yu, F.~Saleh, D.~Campbell, C.~Rodriguez-Opazo, H.~Li, and
  S.~Gould, ``The ikea asm dataset: Understanding people assembling furniture
  through actions, objects and pose,'' 2020.

\bibitem{li2018co}
C.~Li, Q.~Zhong, D.~Xie, and S.~Pu, ``Co-occurrence feature learning from
  skeleton data for action recognition and detection with hierarchical
  aggregation,'' in \emph{Proceedings of the 27th International Joint
  Conference on Artificial Intelligence}.\hskip 1em plus 0.5em minus
  0.4em\relax AAAI Press, 2018, p. 786–792.

\end{thebibliography}
